%% file: main.tex
\documentclass[twoside]{article}

\usepackage[accepted]{aistats2026}

\makeatletter
\renewcommand{\AISTATS@appearing}{%
Proceedings of the 29\textsuperscript{th} International Conference on
Artificial Intelligence and Statistics (AISTATS) 2026,
Tangier, Morocco. PMLR: Volume~300. Copyright 2026 by the author(s). * Equal Contribution%
}
\makeatother

\usepackage{url}
\usepackage{graphicx}    
\usepackage{subcaption}  
\usepackage{multirow}
\usepackage{multicol}
\usepackage{booktabs}
\usepackage{amsfonts}
\usepackage{mathtools}
\usepackage{tcolorbox}   
\usepackage{enumitem}    
\tcbset{fontupper=\footnotesize}  
\newcommand{\modeltrace}[3]{%
  \begin{tcolorbox}[
      title      = #1,
      colframe   = black!60,
      colback    = white!4,
      left=1pt,right=1pt,top=1pt,bottom=1pt,
      boxsep=1pt,          
  ]
  \textbf{Final:} #2

  \vspace{2pt}
  \textbf{Steps:}
  \begin{enumerate}[nosep,leftmargin=*,label=\arabic*.]
    #3
  \end{enumerate}
  \end{tcolorbox}}

\usepackage{listings}
\usepackage{xcolor}
\usepackage[ruled,vlined,linesnumbered]{algorithm2e}
\SetCommentSty{rmfamily}
\usepackage{enumitem}

\newtheorem{thm}{Theorem}[section]
\newtheorem{defn}[thm]{Definition}

\usepackage[round]{natbib}
\usepackage{hyperref}

\begin{document}

%

%

\runningauthor{Wu, Xiong, Li, Yu, et al.}

\newcommand{\authsep}{\hspace{0.9em}}

\twocolumn[
\aistatstitle{CTRLS: Chain-of-Thought Reasoning via Latent State Transition}

\aistatsauthor{
  \makebox[\linewidth][c]{%
    \textbf{Junda Wu}$^{1,*}$\authsep
    \textbf{Yuxin Xiong}$^{1,*}$\authsep
    \textbf{Xintong Li}$^{1}$\authsep
    \textbf{Sheldon Yu}$^{1}$

  }\\[0.7pt]
  \makebox[\linewidth][c]{%
    \textbf{Zhengmian Hu}$^{2}$\authsep
    \textbf{Tong Yu}$^{2}$\authsep
    \textbf{Rui Wang}$^{2}$\authsep
    \textbf{Xiang Chen}$^{2}$\authsep
    \textbf{Jingbo Shang}$^{1}$\authsep
    \textbf{Julian McAuley}$^{1}$%
  }\\[6pt] 
}

\aistatsaddress{ 
  \makebox[\linewidth][c]{%
    $^{1}$University of California, San Diego \authsep
    $^{2}$Adobe Research
  }
}
]

\input{content/0_abstract}

\input{content/1_intro}

\input{content/2_related}

\input{content/3_prelim}

\input{content/3.1.formulation}

\input{content/4_method}
\input{content/5_exp}

\input{content/6_analysis}
\input{content/7_conclusion}

\bibliographystyle{plainnat}
\bibliography{main}

\onecolumn
\appendix

\input{content/9_analysis}

\input{content/8_appendix}

\end{document}

%% file: content/0_abstract.tex
\begin{abstract}
Chain-of-thought (CoT) reasoning enables large language models (LLMs) to break down complex problems into explainable intermediate steps, 
significantly enhancing model transparency and performance in reasoning tasks. 
However, conventional CoT methods rely on heuristic sampling without structured modelling of reasoning transitions, 
constraining their ability to explore and discover diverse and effective reasoning trajectories. 
In this work, we introduce \texttt{CTRLS}, a framework that formulates CoT reasoning as a Markov decision process (MDP) with latent state transitions, 
enabling explainable and state-aware exploration via distributional reinforcement learning. 
By modelling reasoning actions as explicit probability distributions in latent space,
our approach explicitly models epistemic uncertainty, facilitating robust exploration of the reasoning space. 
Enabled by our formulation, we propose an on-policy reinforcement learning scheme to iteratively refine latent transitions without fine-tuning of the underlying LLM.
Theoretical analyses provide evidence lower bounds (ELBO), theoretically grounding our transition-aware modelling of latent reasoning dynamics. 

\end{abstract}

%% file: content/1_intro.tex
\section{Introduction}
Chain-of-thought (CoT) reasoning has emerged as an effective paradigm for enabling large language models (LLMs) to tackle complex tasks by decomposing them into structured,
interpretable intermediate reasoning steps \citep{wei2022chain, kojima2022large,wu2024decot}.
However, conventional CoT prompting lacks transition-aware modelling, such that each step is generated autoregressively without capturing the underlying dynamics of reasoning transitions,
limiting explainable exploration and diversity \cite{yu2025explainable, zhang2025softthink, gan2025rethinking, hou2023mechanistic}, in reasoning trajectories \citep{wei2022chain,kveton2025active,xia2025selection}.
On the other end, structured reasoning frameworks (e.g., Toolchain* \citep{zhuangtoolchain}, program synthesis \citep{zhangplanning}, 
knowledge graph \citep{wuocean}) enforce rigid intermediate steps via API calls or logic traces,
offering structural control but sacrificing flexibility and semantic adaptability \citep{kojima2022large,wuocean}.


\begin{figure*}[t]
    \centering
    \includegraphics[width=.9\linewidth, trim=0 130 0 20, clip]{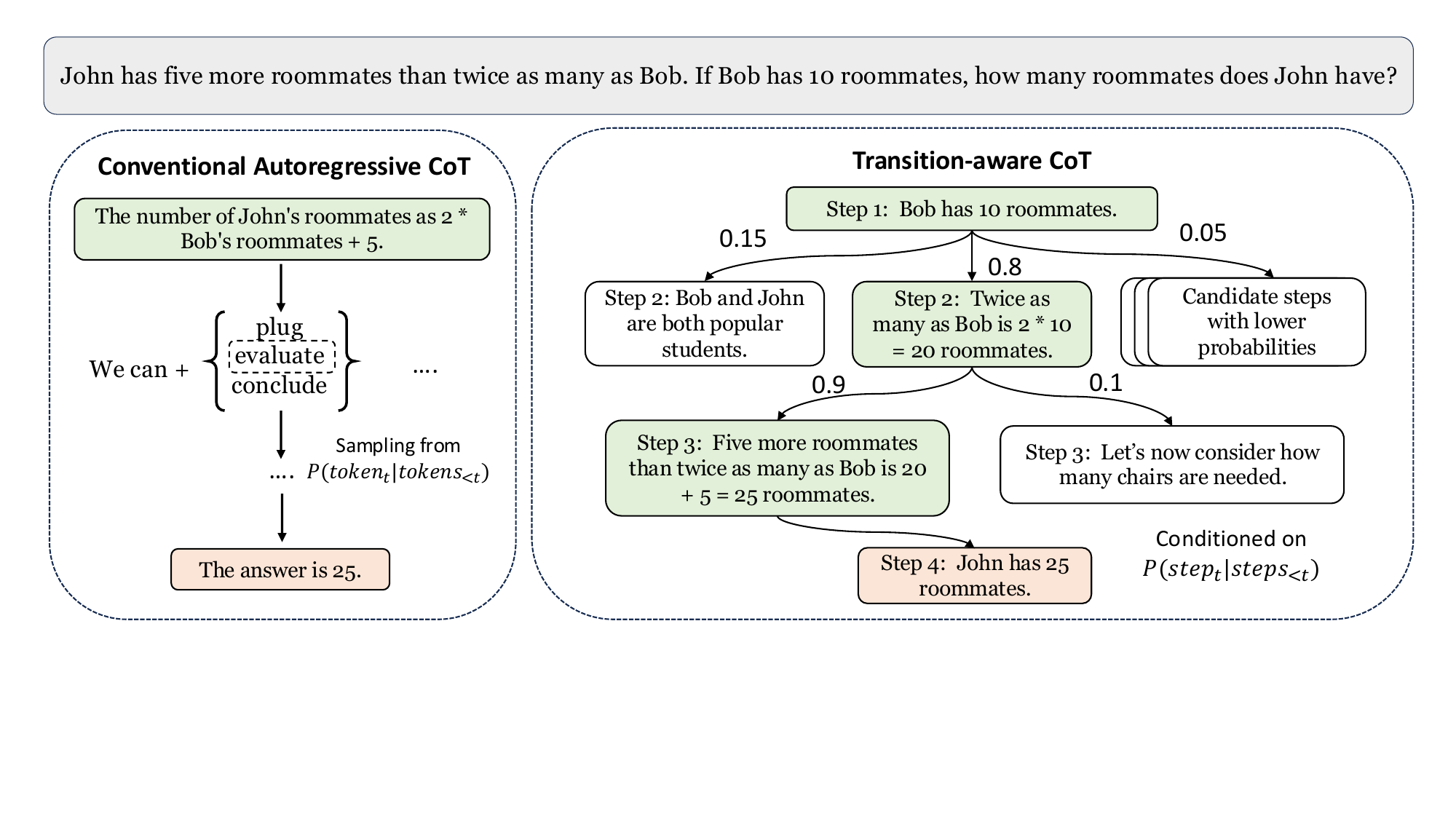}
    \caption{Illustration of the difference between conventional CoT prompting and CTRLS.}
    \label{fig:intro_comparison}
    \vspace{-1em}
\end{figure*}

To illustrate the limitations of conventional CoT prompting, Figure~\ref{fig:intro_comparison} contrasts standard autoregressive reasoning with our transition-aware framework.
On the left, traditional CoT unfolds step-by-step without modelling transitions between reasoning steps, often leading to premature commitment and limited trajectory diversity. 
In contrast, our method models reasoning as a stochastic trajectory over latent states, enabling explicit transition modelling and exploration of alternative reasoning paths.
This comparison highlights the potential of transition-aware CoT to uncover more effective and diverse reasoning strategies.

To bridge the extremes, we propose transition-aware CoT reasoning \texttt{CTRLS}, 
a novel perspective that frames reasoning as structured trajectories within a latent state space. 
Each reasoning step corresponds to a continuous latent semantic state, with transitions dynamically learned via a latent-state MDP formulation. 
This modelling explicitly captures reasoning regularities and semantic relationships between intermediate steps, supporting explainable exploration.

Modelling transition-aware CoT reasoning presents fundamental challenges: 
(i) the need to infer latent semantic states that capture the progression of reasoning steps, despite the lack of explicit supervision; 
(ii) learning stable and generalizable transition dynamics across these latent states; 
and (iii) adapting LLMs to generate coherent reasoning steps conditioned on these latent abstractions. 
To address these challenges holistically, we propose a unified variational model that jointly learns a latent encoder, a transition policy, and a reasoning adapter for the LLM, 
optimized under a single \textbf{evidence lower bound (ELBO)} objective.
This structured design enables semantic grounding and modular reasoning, and naturally supports distributional reinforcement learning \citep{bellemare2017distributional, dabney2018implicit} by treating reasoning actions as stochastic policies over latent states. 
To ensure robust exploration and avoid degenerate trajectories, CTRLS incorporates on-policy optimization with entropy regularization and epsilon-greedy sampling.
Overall, our framework provides a principled and tractable foundation for controllable, structured, and adaptive CoT reasoning.
We summarize our contributions as follows:

\begin{itemize}
\item We introduce a latent-space MDP formulation for explicit modelling of chain-of-thought transitions.
\item We propose a distributional reinforcement learning method tailored for robust and explainable exploration in CoT generation.
\item We theoretically derive finite-sample evidence lower bounds (ELBO) as the learning objective of CTRLS pre-training.
\item We demonstrate empirical improvements in reasoning accuracy, diversity, and exploration efficiency, and showcase enhanced explainability on standard benchmarks. 
\end{itemize}

%% file: content/2_related.tex
\section{Related Work}


\noindent\textbf{Chain-of-Thought Reasoning.}
Chain-of-Thought (CoT) prompting improves large language models (LLMs) by encouraging intermediate reasoning steps before producing final answers~\citep{wei2022chain, chu2023navigate,wu2024decot}. To enhance reasoning quality, prior works introduce self-evaluation~\citep{ling2023deductive,shinn2023reflexion} or integrate external knowledge~\citep{zhao2023verify}. Coconut~\citep{hao2024training} explores reasoning in continuous latent space to bypass token-level constraints. In contrast, we focus on controllability and interpretability via structured latent transitions in language space. Beyond linear CoT, recent work explores structured reasoning, including Tree-of-Thought (ToT)~\citep{yao2023tree} and Chain-of-Preference Optimization (CPO)~\citep{zhang2024chain}, which guide CoT using preferred ToT trajectories. Building on these insights, we model reasoning as transitions over a latent state space, enabling structured step-wise guidance without explicit search.




\noindent\textbf{Reinforcement Learning for LLM Reasoning.}
Reinforcement learning (RL) has been widely used to enhance the reasoning abilities of large language models (LLMs), 
particularly via Reinforcement Learning from Human Feedback (RLHF)~\citep{ouyang2022training, bai2022training} and Direct Preference Optimization (DPO)~\citep{rafailov2023direct,wu2025context},
which learns reward models from human preferences and optimizes LLMs using policy gradient methods such as PPO. 
While effective for preference alignment, RL-based reasoning faces challenges from sparse and delayed rewards. 
To address this, recent works introduce outcome-based rewards~\citep{cobbe2021training, uesato2022solving} or process-based feedback on intermediate steps~\citep{lightman2023let,choudhury2025process}, sometimes leveraging verifiers \citep{su2025expanding,mroueh2025reinforcement,yue2025does} (RLVR) distilled from GPT-4~\citep{zhang2024chain}.
In contrast, we directly model reasoning dynamics in a latent state space and fine-tune transition behaviours using policy gradients, 
without relying on external reward models. This enables scalable and interpretable optimization of reasoning trajectories in a self-contained framework.

%% file: content/3_prelim.tex
\section{Preliminary}


\subsection{Chain-of-thought Reasoning}
Chain-of-thought (CoT) reasoning sequentially generates step-by-step intermediate reasoning toward solving complex tasks by large language models (LLMs)~\citep{wei2022chain,kojima2022large}. 
Given an initial prompt or query $x_0$, the LLM policy $\mu$ ,
iteratively generates reasoning steps $x = (x_1, x_2, \dots, x_T)$, culminating in a final answer prediction $y$. 
Each reasoning step $x_t$ consists of sequentially sampled tokens conditioned on all previously generated tokens:
\begin{equation}
    x_t \sim \mu(\cdot | x_0, x_{<t}), \quad y \sim \mu(\cdot| x_0, x),
\end{equation}


where $x_0$ is the input query or prompt \citep{wei2022chain,wuocean,wu2024commit}.
This autoregressive nature of CoT reasoning presents inherent challenges for controllable generation, as decisions at each step significantly impact subsequent reasoning trajectories.
Instead of explicitly modelling this sequential token generation process, we consider reasoning via latent state-transition,
where we assume a latent state encoded by an \textit{thought abstraction} model $S_t=\rho_\phi(x_{<t})$ and a state transition process $p_\theta(S_{t+1}|S_t)$.

Latent spaces offer a path to more \emph{explainable} decision making. We take a task-driven view in which intermediate steps are semantically meaningful and form transparent trajectories. Following \cite{yu2025explainable, zhang2025softthink, gan2025rethinking, hou2023mechanistic} on steps to human rationale alignment, we further focus on \emph{transition dynamics} among latent reasoning states. Explicitly modeling these transitions makes the process explainable and aligns with our formulation of state-based reasoning and controllable exploration (detailed in Section~\ref{sec:sec4}).



\subsection{Distributional Reinforcement Learning}
Distributional Reinforcement Learning (DRL) explicitly models uncertainty by representing actions as probability distributions rather than deterministic or discrete selections~\citep{bellemare2017distributional,dabney2018implicit}. 
Specifically, we parametrise the policy $\pi_\theta$ to output Dirichlet distributions over potential next reasoning step in CoTs, which naturally express uncertainty and enable richer exploratory behaviour~\citep{chou2017improving}. 
Given state $s_t$, action distributions $a_t$ are drawn as:
\begin{equation}
    a_t \sim \pi_\theta(\cdot|s_t), \quad a_t \in \Delta(\mathcal{A}).
\end{equation}
This formulation inherently captures the uncertainty of stochastic actions over the transition probability distribution of a chain of thoughts. 
The distributional Bellman operator explicitly incorporates uncertainty into state transitions:
\begin{equation}\label{eq:bellman}
\begin{aligned}
    \mathcal{T}Z(s_t, a_t) 
    &\stackrel{D}{=} R(s_t, a_t) + \gamma Z(s_{t+1}, a_{t+1}), \\
    &\quad a_{t+1}\sim \pi_\theta(\cdot|s_{t+1})
\end{aligned}
\end{equation}
where $\stackrel{D}{=}$ denotes distributional equality and $Z(s,a)$ is the return distribution from state-action pairs. 
Thus, we could enable efficient exploration over CoT transition distribution, essential for improved controllability and explainability in downstream tasks~\citep{haarnoja2018soft,lillicrap2015continuous}.

%% file: content/3.1.formulation.tex
\section{Formulation: Transition-aware Chain-of-Thought as an MDP} \label{sec:sec4}
We cast the chain-of-thought (CoT) decoding process into the standard Markov Decision Process (MDP) framework \citep{baionline,zhang2025direct,wuocean}, 
defined by the tuple $(\mathcal{S},\mathcal{A},P,R,\gamma)$. 
This allows us to leverage reinforcement learning algorithms to optimise reasoning trajectories in LLMs.

\noindent\textbf{State and Transition Dynamics}\label{sec:mdp-state}
At each time step $t$, the agent observes a latent reasoning state $s_{t} \in \mathcal{S}$ that captures the semantic context of the prompt and all previously generated reasoning steps.  
Formally, we obtain $s_{t} \sim \rho_{\phi}(\cdot\mid x_0, x_{<t})$ via a stochastic abstraction model $\rho_{\phi}:\mathcal{X}_{<t}\to\Delta(\mathcal{S})$, 
where $\mathcal{X}$ is the space of initial prompts and the sequence of reasoning prefixes (detailed in Section~\ref{sec:state-abstract}).
Here, $x_0$ is the input query or prompt and $x_{<t}=(x_{1},\dots,x_{t-1})$ are the previously generated reasoning segments.
We model latent state evolution with a learned transition matrix $P(s_{t+1}\mid s_{t}, a_{t})$.  
This kernel encapsulates how one reasoning segment influences subsequent latent abstractions and can be trained jointly with the policy.

\noindent\textbf{Distributional Action Sampling}
Our formulation distinctly leverages a distributional perspective for action sampling within the action space $\mathcal{A}$, 
which encompasses distributions over admissible reasoning segments rather than discrete or deterministic selections. 
At each latent reasoning state $s_{t}$, the policy $\pi_{\theta}$ explicitly outputs a probability distribution, 
capturing the inherent epistemic uncertainty in selecting reasoning segments.
This distributional approach facilitates richer exploratory behaviour by enabling the policy to represent a spectrum of plausible reasoning steps, 
each sampled according to a learned Dirichlet distribution:
\begin{equation*}
a_{t} \sim \pi_{\theta}(\cdot\mid s_{t}), \quad a_t \in \Delta(\mathcal{A}), \quad x_{t} \sim P_{\omega}(\cdot \mid a_t, x_{<t}),
\end{equation*}
where $\Delta(\mathcal{A})$ denotes the simplex representing the space of action distributions.
Here, $\theta$ parametrises a policy network designed to output Dirichlet parameters, thus explicitly modelling uncertainty and providing principled exploration strategies in the CoT action space.
$P_{\omega}$ denotes the adapted LLM with adapted parameters introduced in the backbone LLM $\mu$ for action-conditioned generation, which is explained in detail in Section~\ref{sec:condition-cot}. 

\noindent\textbf{Trajectory-level Reward}\label{sec:trajectory_reward}
Aligning with the formulation of LLMs, the quality of the CoT $\tau=(x_{1},\dots,x_{T})$ is evaluated by the final answer $y$. 
Let $y^{\star}$ denote the ground-truth answer associated with the query $x_0$. 
The episodic reward is therefore binary $R(\tau,x_0) = \textbf{1}\{y=y^*\}$.
Our learning objective is to maximise the expected terminal accuracy under the controllable policy
\begin{equation}\label{eq:objective}
J(\theta)=\mathbb E_{\tau\sim\pi_{\theta}(\cdot\mid x_0)}\bigl[ R(\tau,x_0) \bigr],
\end{equation}
where $R(\tau)$ is sparse and unbiased, corresponding to the accuracy reported in the evaluation.

%% file: content/4_method.tex
\section{CTRLS: Chain-of-Thought Reasoning via Latent State Transition}

\begin{figure*}[t]
    \centering
    \includegraphics[width=.8\linewidth]{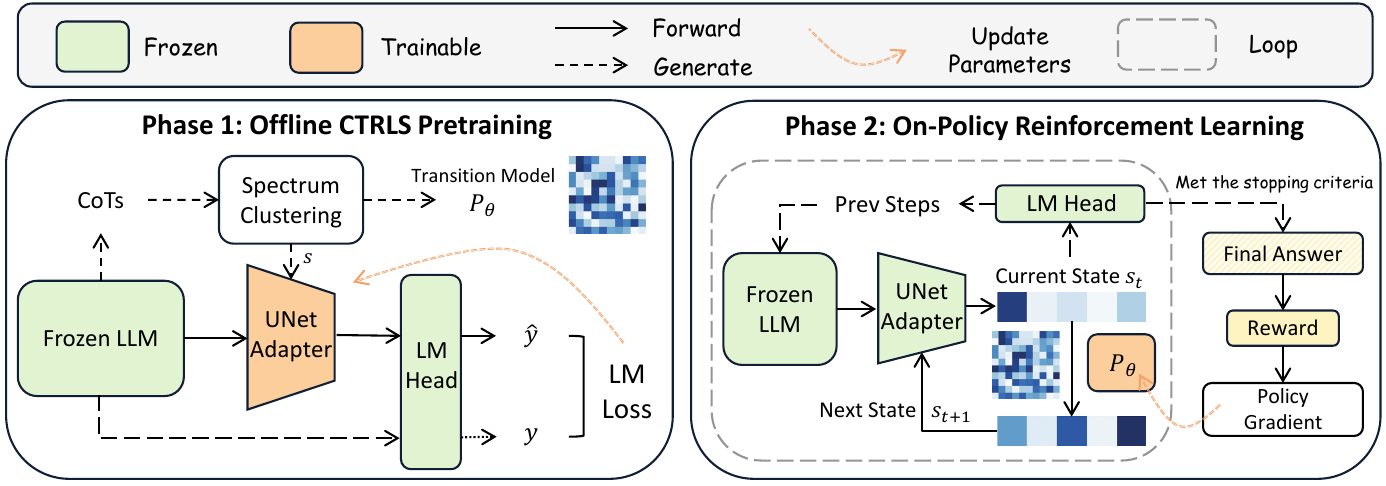}
    \caption{An overview of the proposed two-phase alignment and fine-tuning scheme.}
    \label{fig:overview}
\end{figure*}

Modeling transition-aware CoT reasoning presents unique challenges: 
it requires learning latent state representations to capture the semantic progression of reasoning steps, 
modeling transitions that reflect meaningful and generalizable reasoning dynamics, 
and conditioning the backbone LLM to generate coherent next steps guided by these latent states. 
To address these, CTRLS adopts a unified variational framework and an MDP perspective, implementing three core components:
(i) a stochastic encoder that abstracts reasoning into latent states via an inference model (Section~\ref{sec:state-abstract}), 
(ii) a state-conditioned UNet that injects latent guidance into token representations and models transitions through a policy network (Section~\ref{sec:condition-cot}), 
and (iii) a two-phase alignment and fine-tuning scheme that combines online pre-training with on-policy reinforcement learning (Section~\ref{sec:on-policy}) (illustrated in Figure~\ref{fig:overview}).
This structure supports transition-aware reasoning by optimizing a unified ELBO objective, explicitly modeling uncertainty for principled exploration and controllable, structured generation.

\subsection{Mathematical Assumptions}
\textbf{Assumption 5.1 First-order Markov Assumption:}
The latent reasoning states follow a first-order Markov process. That is, the transition probability depends only on the previous latent state:
\begin{equation*}
    P_\theta(z_t \mid x_{<t}, z_{<t}) = P_\theta(s_t \mid s_{t-1})
\end{equation*}
This assumption simplifies the transition dynamics and is justified by the fact that each latent state encodes the full reasoning prefix.

\textbf{Assumption 5.2 Autoregressive Factorization of the Variational Posterior:}
The variational posterior over latent states is factorized autoregressively to align with the left-to-right generation of LLMs:
\begin{equation*}
    Q_\phi(z_{1:T} \mid x_{1:T}) = \prod_{t=1}^{T} Q_\phi(z_t \mid x_{\leq t})
\end{equation*}
This ensures consistency between the inference model and the sequential nature of language generation.

\subsection{Latent State Encoding}\label{sec:state-abstract}
Based on our MDP formulation in Section~\ref{sec:mdp-state},
we introduce a variational approximation $Q_{\phi}(z_{1:T}|x_{1:T})$, parametrised by $\phi$,
encoding the prompt $x_0$ and reasoning steps $x_{<t}$ into latent state representations. 

\begin{defn}\upshape\textbf{Inference Model via Variational Posterior}
We introduce a variational approximation $Q_{\phi}(z_{1:T}\mid x_{1:T})$.  
Consistent with the sequential nature of the data, it factorises autoregressively:
\begin{equation*}
    Q_{\phi}(z_{1:T}\mid x_{1:T}) = \prod_{t=1}^{T} Q_{\phi}\bigl(z_{t}\mid x_{\le t}\bigr).
\end{equation*}
For each time step $t$, the conditional model $Q_{\phi}\bigl(z_{t}\mid x_{\le t}\bigr)$ outputs a probability distribution over the $N$ possible values of $z_{t}$. 
It can be instantiated by a Gaussian‑mixture posterior, a linear classifier, or a multilayer perceptron (MLP) applied to the hidden representation produced by an encoder.
\end{defn}

The autoregressive factorization mirrors the temporal ordering of the observations, enabling the posterior at step $t$ to depend only on the current and past inputs $x_{\le t}$.  
By sampling $z_{t}$ rather than deterministically encoding it, the model captures both semantic content and epistemic uncertainty, 
reflecting variability in how the system “thinks’’ before emitting each step.  

To instantiate this, we follow~\citep{kveton2025active} to extract token representation $E_t \in \mathbb{R}^{n_t \times d}$ and compute the Gram matrix $G_t = E_t^\top E_t$.
Then, to capture the reasoning semantics, we perform spectrum decomposition and take the flattened most informative top $k$ eigenspaces as the representation,
\begin{equation} \label{eq:spectrum_embedding}
    e_t := \bigl[\sqrt{\lambda_1} \cdot \mathbf{q}_1; \dots; \sqrt{\lambda_k} \cdot \mathbf{q}_k\bigr] \in \mathbb{R}^{kd},
\end{equation}
where $\mathbf{q}_k$ are the corresponding eigenvectors.
We apply $k$-means clustering to all such $e_t$ and each reasoning state  $s_{t-1} \sim \rho_{\phi}(\cdot\mid x_{<t})$ is then assigned as the probability distribution corresponding to clustering centroids $\{\gamma_j\}_{j=1}^K$,
such that $ z_{t} = \sum_{j=1}^{K} s_{t,j} \cdot \gamma_j$.
This continuous relaxation $Q_\phi$ enables structured reasoning over the latent state space and supports subsequent transition modelling.

\subsection{State-aware Chain-of-thought Alignment}\label{sec:condition-cot}

\begin{defn}\upshape\textbf{State-aware Generative Model}
Let $\{(x_t,z_t)\}_{t=1}^{T}$ denote a sequence of observed variables $x_{1:T}\in\mathcal{X}^{T}$ and latent states $z_{1:T}\in\mathcal{Z}^{T}$.  
For parameters $\omega$ (emission) and $\theta$ (transition), the joint distribution factorises autoregressively as
\begin{equation*}
    P_{\omega,\theta}(x_{1:T},z_{1:T})=\prod_{t=1}^{T} P_{\omega,\theta}\bigl(x_t,z_t \mid x_{<t},z_{<t}\bigr),
\end{equation*}
where each factor decomposes into a transition term and an emission term
\begin{equation}\label{eq:inference}
\begin{aligned}
P_{\omega,\theta}\bigl(x_t,z_t \mid x_{<t},z_{<t}\bigr)
&= P_{\theta}\bigl(z_t \mid x_{<t},z_{<t}\bigr)\\
&\enspace \enspace \; P_{\omega}\bigl(x_t \mid x_{<t},z_{\le t}\bigr).
\end{aligned}
\end{equation}
\end{defn}

The transition model $P_{\theta}$ places a prior on the next latent state; it may depend on past observations but is often simplified by the first‑order Markov assumption $P_{\theta}(z_t \mid z_{t-1})$.  
The emission model $P_{\omega}$ generates the current observation conditioned on the complete latent trajectory up to time $t$.  
Together, $P_{\theta}$ and $P_{\omega}$ fully specify the stochastic process underlying the data.

\noindent\textbf{Transition Model $P_{\theta}$}
To model the conditional latent transition, we first encode the previous reasoning steps into a latent state distribution $s_{t-1} \sim \rho_{\phi}(\cdot \mid x_{<t})$, 
as described in Section~\ref{sec:state-abstract}.
Given that the latent representation $z_t$ is deterministically constructed via $z_t = \sum_{j=1}^{K} s_{t,j}\gamma_j$ (see Section~\ref{sec:state-abstract}), 
the transition probability naturally reduces from transitions between latent states $z_t$ to transitions between their underlying distributions $s_t$. 
Specifically, applying the Markov property, we have
\begin{equation}
P_{\theta}(z_t \mid x_{<t}, z_{<t})
= P_{\theta}(s_t \mid s_{<t})
= P_{\theta}(s_t \mid s_{t-1}),
\end{equation}
where the first equality leverages the deterministic relationship between $z_t$ and $s_t$, and the second equality explicitly invokes the Markov assumption~\citep{wu2022dynamics,eysenbach2020off}.

\noindent\textbf{Generation Modelling $P_{\omega}$}
To enable state-aware LLM generation, we design a transformation module that reshapes each token's last hidden state representation $E_{t,i} \in \mathbb{R}^d$ 
conditioned on a step-wise latent reasoning state $s_t \in \mathbb{R}^{d'}$ based on MDP transition. 
Specifically, according to the Markov property
\begin{equation} \label{eq:unet_injection}
\begin{aligned}
    &E'_{t,i} = \mathcal{F}_{\omega_u}\left(\left[\mathcal{F}_{\omega_d}(E_{t,i}); z_t \right]\right), \quad i=1,2,\cdots, n_t\\
    &P_{\omega}(x_t \mid z_{<t}, x_{<t}) = P_{\omega}(x_t \mid z_{t-1}, x_{<t}) \\
    &= \mu(x_t\mid [E'_{1,i<n_1} ; \cdots ; E'_{t-1,i<n_{t-1}}], \omega),
\end{aligned}
\end{equation}
where $\omega=[\omega_d;\omega_u]$ is a U-Net module that encodes token representation $E_{t,i}$ into a low-rank latent, projects and fuses $z_t$ via a bottleneck interaction, 
and decodes the result back to dimension $d$. 
This conditional encoder allows the model to dynamically adjust token representations in sync with the evolving state, with only $O(r(d + d'))$ additional parameters.

Thus, the one‐step inference of state-conditional generation in \eqref{eq:inference} is realised as
\begin{align} \label{eq:joint_factorization}
P_{\omega,\theta} & \left(x_t,z_t \mid x_{<t},z_{<t}\right)   \nonumber  \\
&= P_{\omega}\!\left(x_t \mid z_{<t}, x_{<t}\right)\,
P_{\theta}\!\left(z_t \mid x_{<t}, z_{<t}\right) \nonumber \\
&= \mu\!\left(x_t \mid H_{t-1}, \omega\right)\,
  P_{\theta}\!\left(s_t \mid s_{t-1}\right),  \\
H_{t-1} &\coloneqq
  \bigl[E'_{1,\,i<n_1};\,\cdots;\,E'_{t-1,\,i<n_{t-1}}\bigr].
\end{align}

where such factorization makes clear how latent dynamics and token sampling interlock.
Then we formally introduce the evidence lower bound (ELBO) of the proposed variational model as the objective for model pre-training. 


\begin{thm}[Evidence Lower-Bound]\label{thm:seq-elbo}
Consider a latent‐state generative model with joint density
$P_{\omega,\theta}(x_{1:T},z_{1:T})
  =\prod_{t=1}^{T}
      P_{\omega}\bigl(x_t\mid x_{<t},z_{\le t}\bigr)
      P_{\theta}\bigl(z_t\mid x_{<t},z_{<t}\bigr)$,
and let
$Q_{\phi}(z_{1:T}\mid x_{1:T})
  =\prod_{t=1}^{T}Q_{\phi}\bigl(z_t\mid x_{\le t}\bigr)$
be any variational distribution.  
Then, for the learnable parameters $\omega,\theta,\phi$, the marginal log‑likelihood of the observations admits the lower bound (detailed derivations in Appendix~\ref{app:proof})
\begin{multline} \label{eq:elbo}
\mathcal{L}(\omega,\theta,\phi)
= \sum_{t=1}^T \mathbb{E}_{Q_{\phi}(z_{\le t}\mid x_{\le t})}
   \!\left[\log P_{\omega}(x_t \mid x_{<t}, z_{\le t})\right] \\
- \sum_{t=1}^T \mathbb{E}_{Q_{\phi}(z_{<t}\mid x_{\le t-1})}
   \!\Bigl[
      \mathrm{KL}\!\bigl(Q_{\phi}(z_t \mid x_{\le t}) \,\| \\
      P_{\theta}(z_t \mid x_{<t}, z_{<t})\bigr)
   \Bigr]
\end{multline}
Equality holds if and only if $Q_{\phi}(z_{1:T}\mid x_{1:T}) = P_{\theta}(z_{1:T}\mid x_{1:T})$,
i.e. when the variational posterior matches the true posterior.  
Maximising $\mathcal{L}$ therefore constitutes a tractable surrogate
objective whose optimisation w.r.t \(\omega,\theta,\phi\)
simultaneously (i) maximises the data likelihood and  
(ii) minimises the posterior gap.
\end{thm}

Theorem 5.3 shows that the ELBO provides a tractable way to align latent transitions with reasoning steps. In practice, this means our pretraining objective is not only theoretically sound but also effective: as seen on GSM8K and MATH, optimizing under this objective leads to higher exploration accuracy and fewer spurious steps (detailed later in Section~\ref{exp:explore}).
Based on the derived ELBO, we propose an offline CTRLS pre-training method in Algorithm~\ref{alg:offline}.


\subsection{On-Policy Chain-of-thought Reinforcement Learning} \label{sec:on-policy}

After offline pre-training, we further enable on‐policy reinforcement learning that optimises only the state-transition model $P_\theta$ through trajectories generated by the current policy. 
At each decision step $t$, we sample an action distribution $a_t \sim \pi_{\theta}(\cdot\mid s_t)$ conditioned on the current latent state $s_t$, 
and subsequently generate the reasoning segment $x_t$ through the state-conditioned LLM generation $P_\omega$. 
Iteratively repeating this process yields complete trajectories $\tau = {(s_t, a_t, s_{t+1})}_{t=1}^T$ ending with a final predicted answer $y$.

\noindent\textbf{Trajectory-level Reward and Bellman Function}
We evaluate each trajectory $\tau$ based on the correctness of its final answer $y$ against the ground truth $y^*$, providing a binary episodic reward:
\begin{equation}
R(\tau, x_0) = \textbf{1}\{y = y^*\}.
\end{equation}
We adopt the distributional Bellman operator \eqref{eq:bellman} to model uncertainty explicitly in state transitions and returns.

\noindent\textbf{Exploration via Epsilon-Greedy}
To effectively balance exploration and exploitation, we implement epsilon-greedy exploration.
Specifically, the exploration-enabled action distribution is obtained by mixing the learned Dirichlet distribution with a uniform distribution over actions:
\begin{equation}
\tilde{\pi}_\theta(a|s_t) = (1 - \epsilon) \pi _\theta(a|s_t) + \epsilon \text{Uniform}(\mathcal{A}),
\end{equation}
where $\epsilon \in [0,1]$ controls the exploration-exploitation trade-off. 
With probability $\epsilon$, actions are thus sampled uniformly from the action simplex, 
promoting exploration of diverse reasoning segments, while with probability $1-\epsilon$, 
actions follow the learned Dirichlet distribution, ensuring exploitation of promising behaviours.

\noindent\textbf{Entropy Regularization}
To further encourage robust exploration and prevent premature convergence to suboptimal solutions, we employ entropy regularization. 
By adding an entropy-based penalty to the learning objective, the policy maintains diversity in the action distributions, 
effectively exploring the full action space and discovering potentially more rewarding trajectories. The entropy term is defined as:
\begin{equation}
\mathcal{H}(\pi_\theta(s_t)) = -\sum_{a \in \mathcal{A}} \pi_\theta(a|s_t)\log \pi_\theta(a|s_t).
\end{equation}

\noindent\textbf{Overall Learning Objective and Policy Gradient}
Integrating trajectory rewards and exploration strategies, our reinforcement learning objective maximises expected trajectory rewards augmented with entropy-based exploration:
\begin{equation}
J(\theta)=\mathbb E_{\tau\sim\pi_{\theta}(\cdot\mid x_0)}\Bigl[ R(\tau,x_0) + \alpha \sum_{t=1}^{T}\mathcal{H}(\pi_\theta(s_t)) \Bigr],
\end{equation}
where $\alpha$ controls the strength of entropy regularization. Using the REINFORCE estimator, the gradient of this objective with respect to policy parameters $\theta$ is computed as:
\begin{equation}
\label{eq:pg}
\begin{split}
\nabla_{\theta}J(\theta)
&= \mathbb{E}_{\tau \sim \pi_{\theta}}\Bigl[
   \bigl(R(\tau) + 
   \alpha \sum_{t=1}^{T}\mathcal{H}(\pi_\theta(s_t))\bigr)
\\
&\quad\quad\times 
   \sum_{t=1}^{T}\nabla_{\theta}
   \log \pi_{\theta}(a_{t}\mid s_{t})
   \Bigr].
\end{split}
\end{equation}
This gradient formulation aligns with accuracy-based evaluation metrics in CoT benchmarks, guiding the optimization of the distributional state-transition model. 
We further illustrate the on-policy reinforcement learning process in Algorithm~\ref{alg:rl}.






%% file: content/5_exp.tex
\section{Experiments}

\subsection{Experimental Setup}

We conduct experiments on two instruction-tuned language models, LLaMA-3.2-3B-Instruct~\citep{grattafiori2024llama} and Qwen2.5-3B-Instruct~\citep{qwen2.5}. Both models serve as the backbone for integrating our framework without modifying any pretrained weights. We evaluate on two math reasoning benchmarks, GSM8K~\citep{cobbe2021gsm8k} and MATH~\citep{hendrycks2021measuring}, which might require step-by-step CoT generation to solve arithmetic and competition-level mathematics problems, respectively. Implementation details are in Appendix~\ref{app:impl}.



\input{table/exp-llama-exploration}

\subsection{State Transition-aware Exploration} \label{exp:explore}
To assess the controllability of our pre-trained, transition-aware chain-of-thought generator, we compare CTRLS against the corresponding base models (LlaMA3.2 and Qwen2.5). 
In Table~\ref{tab:explore}, we report results for two generation temperatures, $\eta\in\{0.5,0.7\}$, and an $\epsilon$-greedy exploration strategy with $\epsilon\in\{0.1,0.3,0.5\}$.
For each test question and each $(\eta,\epsilon)$ configuration, we sample 20 reasoning trajectories from both the base model and CTRLS. 
We measure \textbf{exploration accuracy (pass@20)} as the fraction of questions for which the correct answer appears in at least one of the 20 samples, 
and \textbf{success rate (Succ.)} as the proportion of samples that yield a valid final answer after chain-of-thought generation. 

Based on the observations in Table~\ref{tab:explore}, CTRLS consistently outperforms its backbone counterparts in both exploration accuracy and success rate across all datasets. 
Although performance varies with the choice of $\epsilon$, CTRLS surpasses purely temperature-based sampling in every setting, 
confirming that explicit state-transition guidance yields more effective exploratory behaviour.
Such effective control over latent space exploration is consistent with the theoretical guarantees established in Theorem~\ref{thm:seq-elbo}.
The more informative trajectories produced by CTRLS provide a stronger learning signal for subsequent reinforcement learning, 
accelerating policy improvement in later training stages.


\subsection{Impact of Exploration Configurations}\label{sec:exploration}
In Table~\ref{tab:rl_explore_combined}, we experiment on different exploration configurations and study the effects for on-policy reinforcement learning.
\noindent\textbf{Entropy regularisation.}
We observe that raising the entropy weight from $H{=}0$ to $H{=}0.01$ consistently broadens the search space.
For \textbf{LLaMA-3.2}, this translates into a modest but steady gain on both datasets, while \textbf{Qwen2.5} shows an even clearer trend, where $H{=}0.01$ delivers the best accuracy on both datasets, with a minor loss in success rate.
These results confirm that a small amount of entropy pressure prevents premature convergence to high-probability but sub-optimal transitions.

\input{table/exp-ent-epsilon}

\noindent\textbf{$\epsilon$-greedy sampling.}
Injecting stochastic jumps at every step also fosters diversity, but the magnitude of $\epsilon$ matters.
For \textbf{LLaMA-3.2}, a mild setting ($\epsilon{=}0.1$) yields the highest overall gains, while a larger perturbation ($\epsilon{=}0.3$) instead hurts both accuracy and solution validity, echoing the classic explore–exploit trade-off.
\textbf{Qwen2.5} is more robust for both $\epsilon$ values, yet excessive randomisation still reduces the success rate on the harder MATH dataset.

\input{figure/case1}

%% file: table/exp-llama-exploration.tex
{\setlength{\tabcolsep}{4.5pt}
\begin{table*}[htp]
\caption{Comparison of exploration accuracy and success rate of CTRLS and the base models.}
\label{tab:explore}
\small
\centering
\begin{tabular}{c|l|cccccccc}
\toprule
\multirow{3}{*}{Generation Temperature}     & 
\multirow{3}{*}{Exploration} & 
\multicolumn{4}{c}{LlaMA3.2} &
\multicolumn{4}{c}{Qwen2.5} \\
&&\multicolumn{2}{c}{GSM8K} & \multicolumn{2}{c}{MATH}  & \multicolumn{2}{c}{GSM8K} & \multicolumn{2}{c}{MATH} \\ 

\cmidrule(lr){3-4} \cmidrule(lr){5-6} \cmidrule(lr){7-8}  \cmidrule(lr){9-10}

\multicolumn{1}{l|}{}       &                   & pass@20 & Succ. & pass@20 & Succ. & pass@20 & Succ. & pass@20 & Succ. \\ 
\midrule
\multirow{4}{*}{$\eta=0.5$} & Base              & 72.5 & 95.0 & 50.0 & 90.0 & 87.5 & \textbf{100.0} & \textbf{65.0} & 65.0 \\
                            & $\epsilon=0.1$    & 77.5 & \textbf{100.0} & 47.5 & \textbf{95.0} & \textbf{90.0} & \textbf{100.0} & 57.5 & 60.0 \\
                            & $\epsilon=0.3$    & 75.0 & \textbf{100.0} & \textbf{60.0} & \textbf{95.0} & 87.5 & \textbf{100.0} & 62.5 & 52.5 \\
                            & $\epsilon=0.5$    & \textbf{82.5} & \textbf{100.0} & 52.5 & 90.0 & 87.5 & \textbf{100.0} & \textbf{65.0} & \textbf{75.0} \\
\midrule
\multirow{4}{*}{$\eta=0.7$} & Base              & 77.5 & \textbf{100.0} & 47.5 & \textbf{97.5} & 80.0 & \textbf{100.0} & 72.5 & \textbf{67.5} \\
                            & $\epsilon=0.1$    & 77.5 & \textbf{100.0} & 50.0 & \textbf{97.5} & \textbf{85.0} & 97.5 & 62.5 & 65.0 \\ 
                            & $\epsilon=0.3$    & 70.0 & \textbf{100.0} & \textbf{60.0} & \textbf{97.5} & 82.5 & 97.5 & 62.5 & \textbf{67.5} \\
                            & $\epsilon=0.5$    & \textbf{85.0} & \textbf{100.0} & 47.5 & 90.0 & 80.0 & 97.5 & \textbf{77.5} & \textbf{67.5} \\
\bottomrule
\end{tabular}
\end{table*}
}

%% file: table/exp-ent-epsilon.tex
\begin{table*}[ht]
  \centering
  \caption{RL performance under entropy regularization (left block) and $\epsilon$-greedy exploration (right block).}
  \small
  \begin{tabular}{l|l|cc|cc||l|cc|cc}
    \toprule
    \multirow{2}{*}{Model} &
    \multirow{2}{*}{Entropy} &
    \multicolumn{2}{c|}{GSM8K} &
    \multicolumn{2}{c||}{MATH} &
    \multirow{2}{*}{$\epsilon$-Greedy} &
    \multicolumn{2}{c|}{GSM8K} &
    \multicolumn{2}{c}{MATH} \\
    \cmidrule(lr){3-4}\cmidrule(lr){5-6}\cmidrule(lr){8-9}\cmidrule(lr){10-11}
      &  & pass@20 & Succ. & pass@20 & Succ. &  & pass@20 & Succ. & pass@20 & Succ. \\
    \midrule
    \multirow{3}{*}{LlaMA3.2}
      & Base        & 55.6 & 80.0 & 41.0 & 71.8 & Base          & 55.6 & 80.0 & 41.0 & 71.8 \\
      & $H=0.0$   & 57.6 & 80.8 & 42.8 & 72.2 & $\epsilon=0.1$ & 57.4 & 80.4 & 43.0 & 73.4 \\
      & $H=0.01$  & 56.8 & 79.6 & 42.6 & 74.0 & $\epsilon=0.3$ & 55.8 & 79.4 & 42.0 & 71.2 \\
    \midrule
    \multirow{3}{*}{Qwen2.5}
      & Base        & 64.2 & 91.8 & 22.6 & 33.8 & Base          & 64.2 & 91.8 & 22.6 & 33.8 \\
      & $H=0.0$   & 65.8 & 89.6 & 22.8 & 33.4 & $\epsilon=0.1$ & 66.2 & 89.8 & 22.2 & 32.2 \\
      & $H=0.01$  & 66.4 & 90.2 & 23.6 & 34.6 & $\epsilon=0.3$ & 66.0 & 90.2 & 23.4 & 34.2 \\
    \bottomrule
  \end{tabular}
  
  \label{tab:rl_explore_combined}
\end{table*}

%% file: figure/case1.tex
\begin{figure}[ht]
\centering
\begin{tcolorbox}[colframe=blue, colback=blue!5, left=2pt, right=2pt, top=2pt, bottom=2pt, boxsep=0pt]
\small
\textbf{Question}:\;
For how many positive integers $n$ is $n^2-3n+2$ a prime number?\vspace{-1pt}
\end{tcolorbox}

\begin{minipage}[ht]{\linewidth}
  \modeltrace{\;Base (Qwen2.5)}{
\scriptsize
    $\boxed{2}$}{
    \item First, we factorize $n^2 - 3n + 2$. This can be written as $(n-1)(n-2)$.
    \item For $(n-1)(n-2)$ to be a prime number, one of the factors must be 1 and the other must be the prime number itself.
    \item Therefore, for $(n-1)(n-2)$ to be prime, either $n-1 = 1$ or $n-2 = 1$ must hold true.
    \item (\textcolor{red}{Wrong conclusion}) Thus, there are two values of $n$ that satisfy the condition: $n = 2$ and $n = 3$.
  }
\end{minipage}
\vspace{1em}
\begin{minipage}[ht]{\linewidth}
  \modeltrace{\;CTRLS (Qwen2.5)}{
\scriptsize
    $\boxed{1}$}{
    \item (\textcolor{blue}{Same factorization}) First, we factorize $n^2 - 3n + 2$. This can be written as $(n-1)(n-2)$.
    \item (\textcolor{blue}{Same condition}) For $(n-1)(n-2)$ to be a prime number, one of the factors must be equal to 1 because a prime number has exactly two distinct positive divisors: 1 and itself.
    \item (\textcolor{blue}{Same derivation}) Therefore, either $n-1 = 1$ or $n-2 = 1$..
    \item (\textcolor{green}{Self-reflection}) However, we need to check if these values satisfy the condition that $(n-1)(n-2)$ is indeed a prime number.
    \item (\textcolor{green}{reduce wrong answer}) When $n = 2$, $(n-1)(n-2) = (2-1)(2-2) = 1 \cdot 0 = 0$, which is not a prime number.
    \item (\textcolor{green}{Correct conclusion}) Thus, there is only one value of $n$ that satisfies the condition, namely $n = 3$.
  }
\end{minipage}

\vspace{-4pt}
\caption{\textbf{Qualitative comparison.}  
CTRLS correctly verifies candidate solutions and filters out invalid cases, while the baseline fails to check whether the resulting value is truly prime.}

\label{fig:qualitative}
\end{figure}

%% file: content/6_analysis.tex
\begin{table}[ht]
\centering
\small
\setlength{\tabcolsep}{6pt} 
\renewcommand{\arraystretch}{0.95} 
\caption{GPT-4 rubric scores (0–10) on GSM8K with Qwen. Averaged over 100 samples per setting. We evaluate on four different aspects:
(\textbf{S})elf-reflection, (\textbf{A})lgebra, (\textbf{C})oherence, (\textbf{H})allucination.
}
\begin{tabular}{lccccc}
\toprule
Setting & S & A & C & H↓ & Overall \\
\midrule
Base & 7.62 & 8.44 & 8.23 & 8.51 & 8.05 \\
$\epsilon=0.0, H=0.0$ & 7.82 & 8.83 & 8.43 & 8.66 & 8.34 \\
$\epsilon=0.1, H=0.0$ & 7.74 & 8.65 & 8.33 & 8.51 & 8.17 \\
$\epsilon=0.1, H=0.01$ & 7.73 & 8.54 & 8.35 & 8.55 & 8.16 \\
$\epsilon=0.3, H=0.0$ & 7.80 & 8.77 & 8.45 & 8.68 & 8.30 \\
$\epsilon=0.3, H=0.01$ & 7.72 & 8.73 & 8.37 & 8.60 & 8.20 \\
\bottomrule
\end{tabular}
\label{tab:gptjudge_qwen_gsm8k}
\end{table}

\subsection{Case Study} \label{sec:analysis}


\noindent\textbf{Self-reflection}\quad
One notable advantage of our transition-based reasoning is its ability to perform self-reflection. 
As illustrated in Figure~\ref{fig:qualitative}, both the baseline and CTRLS derive the correct candidate values for $n$ such that $n^2 - 3n + 2$ is prime. 
However, the baseline prematurely outputs both values as correct without verifying whether the resulting expression is actually a prime number. 
In contrast, CTRLS continues reasoning by explicitly validating the primality condition for each candidate. 
It correctly identifies $n = 2$ as invalid, as $(2 - 1)(2 - 2) = 1 \cdot 0 = 0$, which is not a prime. 
This self-reflective validation step leads to the correct final answer. 
Such state-aware step transitions, which is semantically grounded on explainable states, help avoid early commitment to incorrect conclusions.

\noindent\textbf{Corrected algebra errors and reduced hallucinated steps} \quad
CTRLS enhances symbolic reasoning by reducing common algebraic mistakes—such as misapplied formulas and incorrect substitutions—as well as suppressing spurious steps that lack logical grounding. It preserves variable relationships across steps, avoids unnecessary formalisms, and maintains alignment with the problem’s structure. Representative examples are provided in Appendix~\ref{sec:case}.


\noindent\textbf{LLM-as-a-judge evaluation.} 
To complement accuracy-based metrics, we additionally include a GPT-4 based rubric evaluation. 
For each configuration, we randomly sample 100 model outputs on GSM8K and let GPT-4 score them on a 1–10 scale across five criteria: 
self-reflection, algebra correctness, logical coherence, reduction of hallucinated steps, and overall quality. 
Table~\ref{tab:gptjudge_qwen_gsm8k} summarises the averaged results for Qwen; detailed prompts and evaluation pipeline are deferred to Appendix~\ref{app:gptjudge}.

%% file: content/7_conclusion.tex
\section{Conclusion}

We presented CTRLS, a principled framework for transition-aware chain-of-thought reasoning that casts step-wise generation as a latent-state Markov decision process. By modelling reasoning dynamics through distributional policies and optimizing latent transitions via reinforcement learning, CTRLS enables structured, interpretable, and controllable CoT generation. Our experiments demonstrate that CTRLS consistently improves reasoning accuracy, exploration efficiency, and robustness across math benchmarks, and further showcases explainability. Beyond performance, qualitative analyses highlight its ability to recover from symbolic errors, suppress spurious reasoning, and engage in self-reflective correction. We believe CTRLS offers a foundation for more systematic and verifiable reasoning in large language models.

%% file: content/9_analysis.tex
\section{Evidence Lower Bound (ELBO) Derivation} \label{app:proof}

We aim to maximize the log-likelihood $\log P_{\omega,\theta}(x_{1:T})$. Since direct maximization is intractable due to the summation over $z_{1:T}$, we maximize the Evidence Lower Bound (ELBO), $\mathcal{L}(\omega, \theta, \phi)$.

Starting from the definition of the log-likelihood and introducing the variational distribution $Q_{\phi}(z_{1:T}|x_{1:T})$:
\begin{align*}
\log P_{\omega,\theta}(x_{1:T}) &= \log \sum_{z_{1:T}} P_{\omega,\theta}(x_{1:T}, z_{1:T}) \\
&= \log \sum_{z_{1:T}} Q_{\phi}(z_{1:T}|x_{1:T}) \frac{P_{\omega,\theta}(x_{1:T}, z_{1:T})}{Q_{\phi}(z_{1:T}|x_{1:T})} \\
&\ge \sum_{z_{1:T}} Q_{\phi}(z_{1:T}|x_{1:T}) \log \frac{P_{\omega,\theta}(x_{1:T}, z_{1:T})}{Q_{\phi}(z_{1:T}|x_{1:T})} \quad \text{(Jensen's Inequality)} \\
&= \mathbb{E}_{Q_{\phi}(z_{1:T}|x_{1:T})} \left[ \log \frac{P_{\omega,\theta}(x_{1:T}, z_{1:T})}{Q_{\phi}(z_{1:T}|x_{1:T})} \right] \\
&=: \mathcal{L}(\omega, \theta, \phi)
\end{align*}

Now, we expand the ELBO using the factorizations of $P$ and $Q$:
\begin{align*}
\mathcal{L}(\omega, \theta, \phi) &= \mathbb{E}_{Q_{\phi}(z_{1:T}|x_{1:T})} \left[ \log P_{\omega,\theta}(x_{1:T}, z_{1:T}) - \log Q_{\phi}(z_{1:T}|x_{1:T}) \right] \\
&= \mathbb{E}_{Q_{\phi}(z_{1:T}|x_{1:T})} \left[ \sum_{t=1}^T \log P_{\omega,\theta}(x_t, z_t | x_{<t}, z_{<t}) - \sum_{t=1}^T \log Q_{\phi}(z_t | x_{\leq t}) \right] \\
&= \mathbb{E}_{Q_{\phi}(z_{1:T}|x_{1:T})} \left[ \sum_{t=1}^T \left( \log P_{\omega}(x_t | x_{<t}, z_{\le t}) + \log P_{\theta}(z_t | x_{<t}, z_{<t}) - \log Q_{\phi}(z_t | x_{\leq t}) \right) \right] \\
&= \sum_{t=1}^T \mathbb{E}_{Q_{\phi}(z_{1:T}|x_{1:T})} \left[ \log P_{\omega}(x_t | x_{<t}, z_{\le t}) + \log P_{\theta}(z_t | x_{<t}, z_{<t}) - \log Q_{\phi}(z_t | x_{\leq t}) \right]
\end{align*}
For the term at index $t$, the expectation only needs to be taken over $z_{\le t}$, as the terms do not depend on $z_{>t}$. Let $Q_{\phi}(z_{\le t}|x_{1:T})$ denote the marginal distribution of $z_{\le t}$ under $Q_{\phi}(z_{1:T}|x_{1:T})$. With our factorization $Q_{\phi}(z_{1:T}|x_{1:T}) = \prod_{i=1}^T Q_{\phi}(z_i | x_{\leq i})$, the distribution $Q_{\phi}(z_{\le t}|x_{1:T})$ simplifies to $\prod_{i=1}^t Q_{\phi}(z_i | x_{\leq i})$. We use the shorthand $Q_{\phi}(z_{\le t}|x_{\leq t})$ as in the outline, representing this product distribution.
\begin{align*}
\mathcal{L}(\omega, \theta, \phi) &= \sum_{t=1}^T \mathbb{E}_{Q_{\phi}(z_{\le t}|x_{\leq t})} \left[ \log P_{\omega}(x_t | x_{<t}, z_{\le t}) + \log P_{\theta}(z_t | x_{<t}, z_{<t}) - \log Q_{\phi}(z_t | x_{\leq t}) \right] \\
&= \sum_{t=1}^T \left( \mathbb{E}_{Q_{\phi}(z_{\le t}|x_{\leq t})} \left[ \log P_{\omega}(x_t | x_{<t}, z_{\le t}) \right] + \mathbb{E}_{Q_{\phi}(z_{\le t}|x_{\leq t})} \left[ \log P_{\theta}(z_t | x_{<t}, z_{<t}) - \log Q_{\phi}(z_t | x_{\leq t}) \right] \right)
\end{align*}
Now consider the second expectation term. We can rewrite the expectation over $z_{\le t}$ as an expectation over $z_{<t}$ followed by an expectation over $z_t$:
$\mathbb{E}_{Q_{\phi}(z_{\le t}|x_{\leq t})} [\cdot] = \mathbb{E}_{Q_{\phi}(z_{<t}|x_{\leq t-1})} \left[ \mathbb{E}_{Q_{\phi}(z_t|x_{\leq t})} [\cdot] \right]$.
\begin{align*}
&\mathbb{E}_{Q_{\phi}(z_{\le t}|x_{\leq t})} \left[ \log P_{\theta}(z_t | x_{<t}, z_{<t}) - \log Q_{\phi}(z_t | x_{\leq t}) \right] \\
&= \mathbb{E}_{Q_{\phi}(z_{<t}|x_{\leq t-1})} \left[ \sum_{z_t} Q_{\phi}(z_t | x_{\leq t}) \left( \log P_{\theta}(z_t | x_{<t}, z_{<t}) - \log Q_{\phi}(z_t | x_{\leq t}) \right) \right] \\
&= \mathbb{E}_{Q_{\phi}(z_{<t}|x_{\leq t-1})} \left[ - \sum_{z_t} Q_{\phi}(z_t | x_{\leq t}) \log \frac{Q_{\phi}(z_t | x_{\leq t})}{P_{\theta}(z_t | x_{<t}, z_{<t})} \right] \\
&= - \mathbb{E}_{Q_{\phi}(z_{<t}|x_{\leq t-1})} \left[ \text{KL} \left( Q_{\phi}(z_t | x_{\leq t}) \parallel P_{\theta}(z_t | x_{<t}, z_{<t}) \right) \right]
\end{align*}
Substituting this back into the ELBO expression:
\begin{align*}
\mathcal{L}(\omega, \theta, \phi) = \sum_{t=1}^T \left( \mathbb{E}_{Q_{\phi}(z_{\le t}|x_{\leq t})} \left[ \log P_{\omega}(x_t | x_{<t}, z_{\le t}) \right] - \mathbb{E}_{Q_{\phi}(z_{<t}|x_{\leq t-1})} \left[ \text{KL} \left( Q_{\phi}(z_t | x_{\leq t}) \parallel P_{\theta}(z_t | x_{<t}, z_{<t}) \right) \right] \right)
\end{align*}
This is the final form of the ELBO. It consists of two main terms summed over time:

1.  \textbf{Expected Reconstruction Log-Likelihood}: The expectation of the log-probability of the observed data $x_t$ given the history and the inferred latent states $z_{\le t}$.

2.  \textbf{Expected KL Divergence}: The negative KL divergence between the approximate posterior $Q_{\phi}(z_t | x_{\leq t})$ and the latent prior/transition model $P_{\theta}(z_t | x_{<t}, z_{<t})$, averaged over the inferred previous latent states $z_{<t}$. This term acts as a regularizer, encouraging the approximate posterior to stay close to the prior.

Maximizing this ELBO with respect to $\omega$, $\theta$, and $\phi$ trains the model. The expectations are typically approximated using samples from $Q_{\phi}$.

%% file: content/8_appendix.tex
\section{Algorithm of CTRLS Pre-training}

Algorithm~\ref{alg:offline} outlines the pre-training procedure of CTRLS. The model is trained to align step-wise reasoning with latent state transitions using supervised CoT trajectories and associated state embeddings.

\input{table/algo_training}

\section{Algorithm of CTRLS for On-policy Reinforcement Learning}

Algorithm~\ref{alg:rl} details the on-policy reinforcement learning phase. The model fine-tunes its transition policy using trajectory-level rewards and sampled state dynamics to improve reasoning performance.

\input{table/algo_rl}

\subsection{On-policy Reinforcement Learning}\label{exp:rl}

Building on the exploration scheme, we fine‑tune both backbones on GSM8K and MATH with on‑policy RL under four settings: 
(i) $\epsilon$-greedy only, (ii) entropy regularisation only, (iii) both techniques, and (iv) no exploration (baseline). 
Figure~\ref{fig:llama-rl-gsm8k} shows the on-policy learning curves. 
For LlaMA3.2, entropy regularization is crucial, as the policy collapses to a few high‑probability actions and training stalls when it is disabled. 
CTRLS improves in the early phase (i.e., first 500 steps) without entropy regularisation, but its performance subsequently degrades as action diversity in the transition distribution diminishes.

\input{figure/both-rl}

In addition, we observe that entropy regularization prevents this collapse, and combining it with $\epsilon$-greedy sampling yields the fastest and most stable gains, confirming that the two mechanisms are complementary and that CTRLS benefits most when both are employed.
We observe a similar trend for Qwen2.5 backbone model in Figure~\ref{fig:qwen-rl-gsm8k}, while CTRLS achieves more robust on-policy learning improvement even without entropy regularization.

In Figure~\ref {fig:both-math}, we further present the on-policy learning of both LlaMA3.2 and Qwen2.5 models of CTRLS on MATH dataset.
As discussed in Section~\ref{exp:rl}, the $\epsilon$-greedy exploration and entropy regularization enhance the learning robustness on LlaMA3.2 based CTRLS.
However, we also observe learning degeneration on MATH dataset for Qwen2.5 model. Such problem is derived from the sensitivity of exploration instability discussed in Section~\ref{sec:exploration}.

\input{figure/both-math}

\section{Case Study}
\label{sec:case}

We provide qualitative examples to support the findings discussed in Section~\ref{sec:analysis}. The Figure~\ref{fig:qualitative2} illustrates how CTRLS corrects an algebraic substitution error made by the baseline. The Figure~\ref{fig:qualitative3} shows how CTRLS avoids hallucinated symbolic steps and follows a simpler, task-grounded reasoning path.

\input{figure/case2}

\input{figure/case3}

\section{GPT-as-judge Evaluation Details} \label{app:gptjudge}

We implement an automated evaluation pipeline to quantify qualitative aspects of reasoning.
The pipeline is based on the following steps:

\begin{enumerate}
  \item \textbf{Sampling.} For each experiment configuration, we subsample 100 solutions from the model outputs.
  \item \textbf{Prompt construction.} Each solution is converted into a structured prompt containing:
    the original question, the ground-truth answer, the step-by-step reasoning, and the model's final answer.
  \item \textbf{GPT-4 evaluation.} We query GPT-4 (temperature $0.1$) with instructions to respond strictly in JSON format. 
  The model assigns scores (1–10) for the following criteria:
    \begin{itemize}
      \item Self-reflection and validation
      \item Correctness of algebra
      \item Logical coherence
      \item Reduction of hallucinated steps
      \item Overall quality
    \end{itemize}
  \item \textbf{Post-processing.} We parse the JSON output, compute averages, and report aggregate results. 
  When parsing fails (e.g., malformed JSON), we apply simple regex extraction or fall back to default values.
\end{enumerate}

\section{Implementation Details}\label{app:impl}
We follow the prompt design in~\citep{wei2022chain} to guide step-wise CoT generation. For each question, we sample multiple CoT trajectories and retain only those yielding the correct final answer. The filtered set forms the training data for CTRLS pretraining. Token embeddings are projected into spectrum embeddings, clustered into $K{=}64$ latent states, and used to estimate the transition matrix. To encourage diverse reasoning paths, sampling with temperature and top-$k$ filtering is applied during generation. The entire process takes approximately 2 hours on two A6000 GPUs.

%% file: table/algo_training.tex
\begin{algorithm}[htp]
\caption{Offline CTRLS Pretraining}
\label{alg:offline}
\KwIn{Dataset $\mathcal{D}$; pretrained LLM $P_\omega$; number of clusters $K$}
\KwOut{CTRLS generator $P_\omega$; transition model $P_\theta$; inference model $Q_\phi$}

\For{each $(x, c_{1:T}) \in \mathcal{D}$}{
    \For{each step $c_t$}{
        $E_t \gets$ token embeddings from $P_\omega$\;
        $G_t \gets E_t^\top E_t$ \tcp*[r]{Gram matrix}
        $e_t \gets$ spectrum features of $G_t$ \tcp*[r]{Eq.~(\ref{eq:spectrum_embedding})}
        Save $e_t$\;
    }
}
Cluster $\{e_t\}$ to obtain centroids $\{\gamma_1, ..., \gamma_K\}$\;
Compute soft assignments $\{s_t\}$ for all steps \tcp*[r]{State distributions}
Define $Q_\phi$ as the combination of spectral encoder and soft assignment mechanism \;
Optimize $P_\theta$ to fit transition pairs $(s_{t-1} \rightarrow s_t)$ via KL \tcp*[r]{Eq.~(\ref{eq:elbo})}

\For{each $(x, c_{1:T}) \in \mathcal{D}$}{
    \For{each step $c_t$}{
        $z_t \gets \sum_{j=1}^K s_{t,j} \cdot \gamma_j$ \;
        Inject $z_t$ into token representations \tcp*[r]{Eq.~(\ref{eq:unet_injection})}
    }
    Update $P_\omega$ to minimize supervised loss
}
\Return $P_\omega$, $P_\theta$, $Q_\phi$

\end{algorithm}

%% file: table/algo_rl.tex


\begin{algorithm}[htp]
\caption{On-Policy RL Fine-tuning for CTRLS}
\label{alg:rl}
\KwIn{Pretrained generator $P_\omega$; transition model $P_\theta$; inference model $Q_\phi$; centroids $\{\gamma_1, ..., \gamma_K\}$; training dataset $\mathcal{D}$; number of steps $T$}
\KwOut{Fine-tuned transition model $P_\theta$}

\For{each $(x, y^*) \in \mathcal{D}$}{
    $s_0 \gets Q_\phi(x)$ \tcp*[r]{Initial latent state from prompt}
    Set $x_0 \gets x$; trajectory $\tau \gets \emptyset$\;

    \For{$t = 1$ to $T$}{
        $z_t \gets \sum_j s_{t,j} \cdot \gamma_j$ \tcp*[r]{Weighted latent vector}
        Inject $z_t$ into token embeddings via Eq.~(\ref{eq:unet_injection}) \;
        Sample $x_t \sim P_\omega(\cdot \mid x_{<t}, z_{\le t})$ \tcp*[r]{State-aware generation}
        Append $(s_t, x_t)$ to trajectory $\tau$ \;

        \If{stopping criterion met (e.g., EOS token)}{
            \textbf{break}
        }

        Sample $s_{t+1} \sim \epsilon\text{-Greedy}(P_\theta(\cdot \mid s_t))$ \tcp*[r]{Latent transition (soft dist.)}
    }

    Compute answer $\hat{y}$ from full $x_{1:T}$\;
    Set reward $R = \mathbf{1}\{\hat{y} = y^*\}$ \;
    Compute policy gradient with entropy bonus \;
    Update $P_\theta$ using optimizer \;
}
\Return $P_\theta$
\end{algorithm}

%% file: figure/both-rl.tex
\begin{figure}[htbp]
  \centering
  \begin{subfigure}[t]{0.495\textwidth} 
    \includegraphics[width=\linewidth]{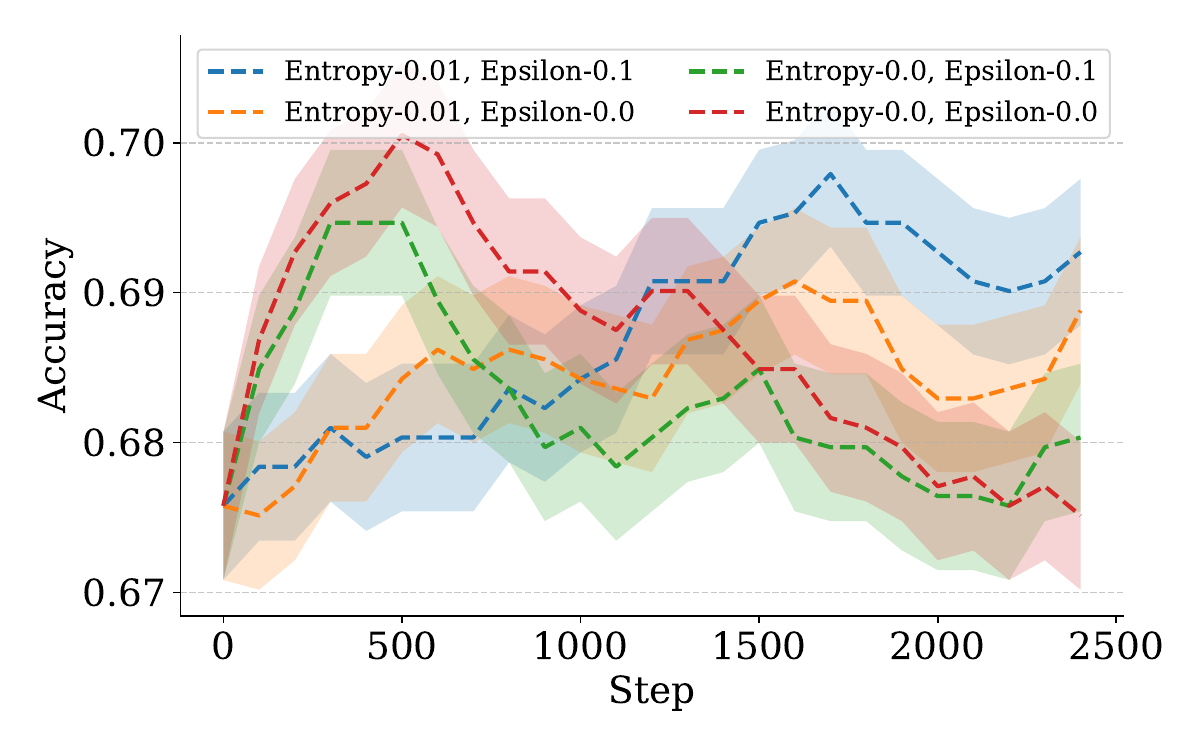}
    \caption{LlaMA3.2}
    \label{fig:llama-rl-gsm8k}
  \end{subfigure}
  \begin{subfigure}[t]{0.49\textwidth}
    \centering
    \includegraphics[width=\linewidth]{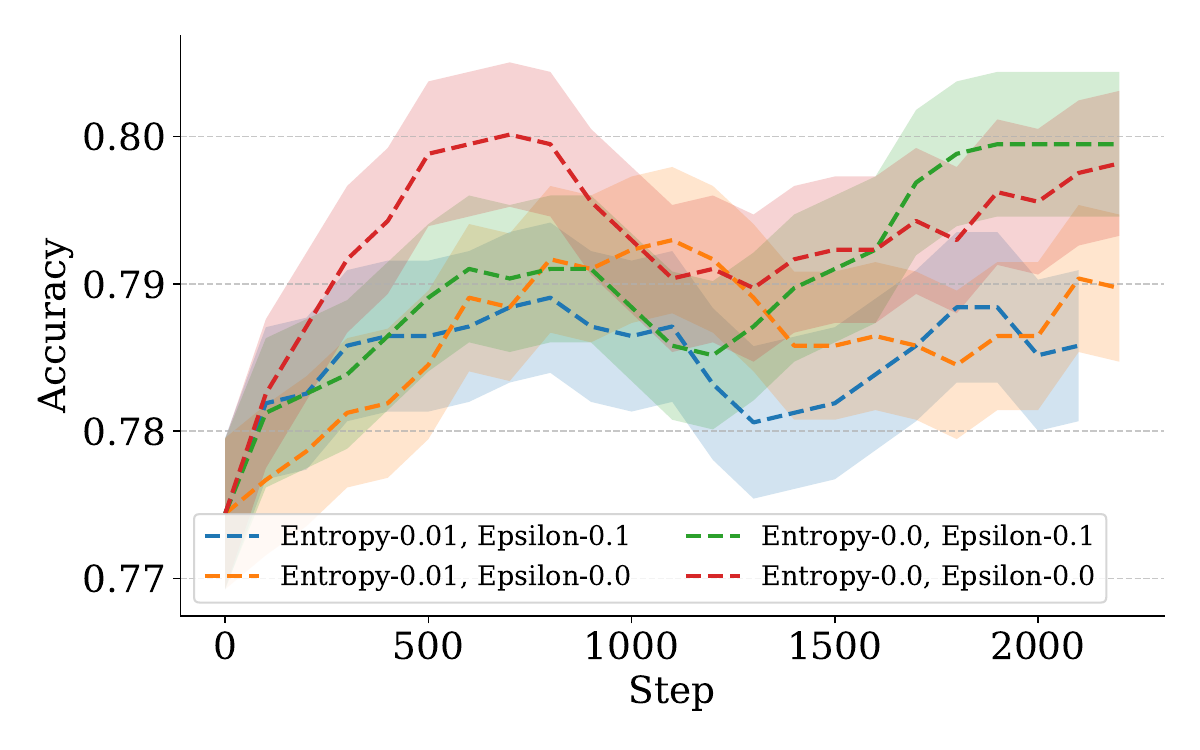}
    \caption{Qwen2.5}
    \label{fig:qwen-rl-gsm8k}
  \end{subfigure}
  
  \caption{On-policy learning curves for CTRLS with \textbf{LlaMA3.2} and \textbf{Qwen2.5}.}
  \label{fig:both-rl}
  \vspace{-1em}
\end{figure}

%% file: figure/both-math.tex
\begin{figure}[htbp]
  \centering
  \begin{subfigure}[t]{0.495\textwidth}
    \includegraphics[width=\linewidth]{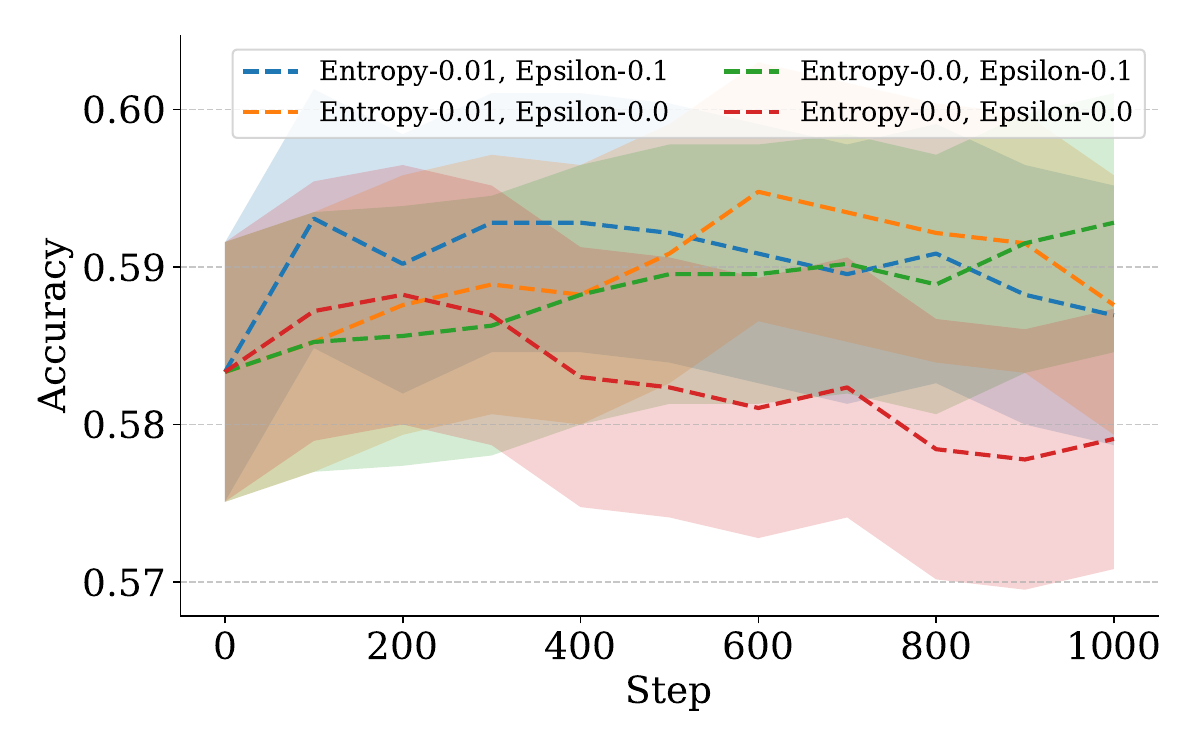}
    \caption{MATH}
    \label{fig:llama-rl-math}
  \end{subfigure}
  \begin{subfigure}[t]{0.49\textwidth}
    \centering
    \includegraphics[width=\linewidth]{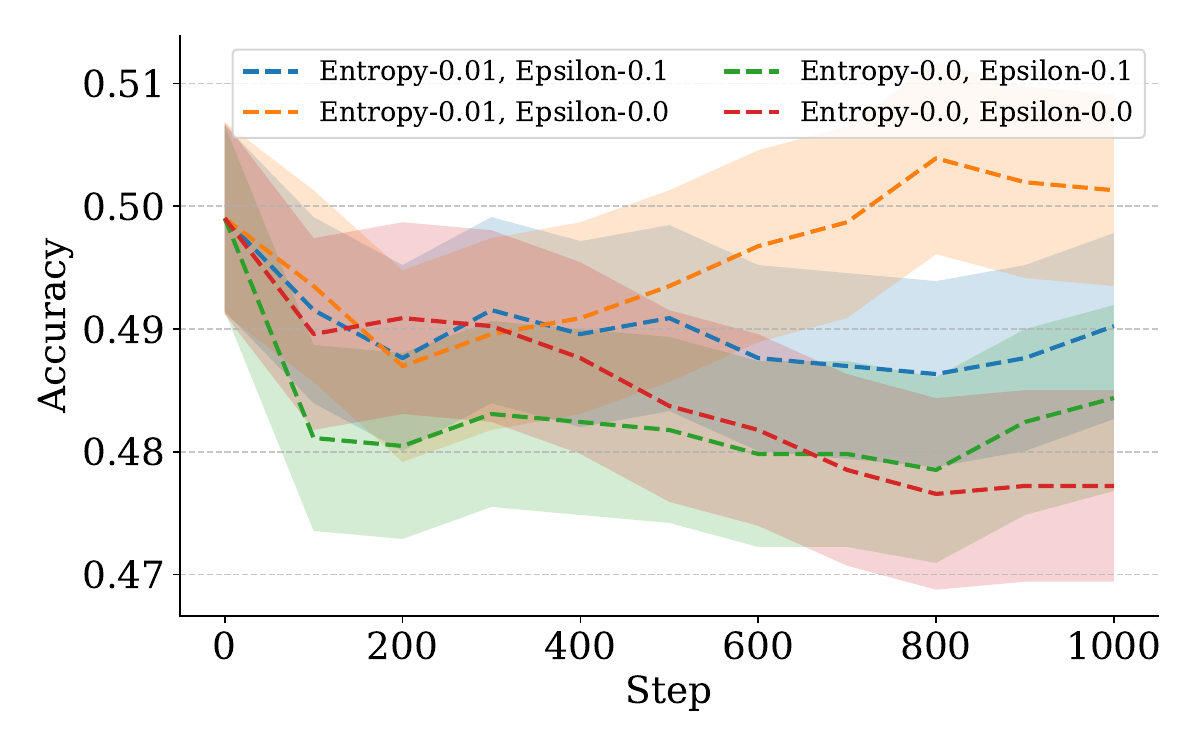}
    \caption{MATH}
    \label{fig:qwen-rl-math}
  \end{subfigure}

  \caption{On-policy reinforcement learning curves for \textbf{LlaMA3.2} and \textbf{Qwen2.5} on MATH dataset.}
  \label{fig:both-math}
\end{figure}

%% file: figure/case2.tex
\begin{figure}[t]
\centering
\begin{tcolorbox}[colframe=blue, colback=blue!5, left=2pt, right=2pt, top=2pt, bottom=2pt, boxsep=0pt]
\small
\textbf{Question}:\;
Let $a_1,a_2,a_3,\dots$ be an arithmetic sequence. If $\frac{a_4}{a_2} = 3$, what is $\frac{a_5}{a_3}$?
\vspace{-1pt}
\end{tcolorbox}

\begin{minipage}[t]{\linewidth}
  \modeltrace{\;Base (LlaMA3.2)}{
\scriptsize
    $\boxed{4}$}{
    \item In an arithmetic sequence, the difference between consecutive terms is constant. Let's denote this common difference as $d$.
    \item We can express $a_4$ as $a_2 + 2d$ and $a_5$ as $a_3 + 2d$. Given that $\frac{a_4}{a_2} = 3$, we can write this as $\frac{a_2 + 2d}{a_2} = 3$.
    \item (\textcolor{red}{Wrong derivation}) Solving for $d$, we get $2d = 3a_2$, which simplifies to $d = \frac{3}{2}a_2$. Now, we can express $a_5$ as $a_3 + 2d = a_3 + 2\left(\frac{3}{2}a_2\right) = a_3 + 3a_2$.
    \item (\textcolor{red}{Wrong conclusion}) Therefore, the ratio $\frac{a_5}{a_3}$ is equal to $\frac{a_3 + 3a_2}{a_3} = 1 + \frac{3a_2}{a_3} = 1 + 3 = 4$.
  }
\end{minipage}
\vspace{1em}
\begin{minipage}[t]{\linewidth}
  \modeltrace{\;CTRLS (LlaMA3.2)}{
\scriptsize
    $\boxed{2}$}{
    \item (\textcolor{blue}{Same setup}) In an arithmetic sequence, the difference between consecutive terms is constant. Let's denote this common difference as $d$.
    \item (\textcolor{blue}{Same derivation}) We can express $a_4$ as $a_2 + 2d$ and $a_5$ as $a_3 + 2d$. Given that $\frac{a_4}{a_2} = 3$, we can write this as $\frac{a_2 + 2d}{a_2} = 3$.
    \item (\textcolor{green}{Correct derivation}) Solving for $d$, we get $2d = 2a_2$, which simplifies to $d = a_2$. Now, we can express $a_5$ as $a_3 + 2a_2$ and $a_3$ as $a_2 + a_2$, which simplifies to $a_3 = 2a_2$.
    \item (\textcolor{green}{Correct conclusion}) Substituting these expressions into $\frac{a_5}{a_3}$, we get $\frac{a_3 + 2a_2}{a_2 + a_2} = \frac{2a_2 + 2a_2}{2a_2} = \frac{4a_2}{2a_2} = 2$.
  }
\end{minipage}

\vspace{-4pt}
\caption{\textbf{Qualitative comparison.} CTRLS corrects a symbolic substitution error made by the baseline in an arithmetic reasoning task, resulting in a more accurate final answer.}
\label{fig:qualitative2}
\end{figure}

%% file: figure/case3.tex
\begin{figure}[t]
\centering
\begin{tcolorbox}[colframe=blue, colback=blue!5, left=2pt, right=2pt, top=2pt, bottom=2pt, boxsep=0pt]
\small
\textbf{Question}:\;
For how many real values of $x$ is $\sqrt{63-\sqrt{x}}$ an integer?
\vspace{-1pt}
\end{tcolorbox}

\begin{minipage}[t]{\linewidth}
  \modeltrace{\;Base (LlaMA3.2)}{
\scriptsize
    $\boxed{4}$}{
    \item We start by letting $n = \sqrt{63-\sqrt{x}}$, which implies that $n^2 = 63 - \sqrt{x}$. This further simplifies to $\sqrt{x} = 63 - n^2$, and squaring both sides gives us $x = (63-n^2)^2$.
    \item (\textcolor{red}{Hallucinated step}) We can rewrite this equation as $n^2 + m^2 = 63$, which is a form of Pell's equation.
    \item (\textcolor{red}{Misled solution}) This gives us the possible values of $n$ as $3, 9, 15$, and $21$, and the corresponding values of $m$ as $6, 0, -6$, and $-12$, respectively.
    \item (\textcolor{red}{Wrong conclusion}) Therefore, there are $\boxed{4}$ possible values of $x$ for which $\sqrt{63-\sqrt{x}}$ is an integer.
  }
\end{minipage}
\vspace{1em}
\begin{minipage}[t]{\linewidth}
  \modeltrace{\;CTRLS (LlaMA3.2)}{
\scriptsize
    $\boxed{8}$}{
    \item (\textcolor{blue}{Same setup}) We start by letting $n = \sqrt{63-\sqrt{x}}$, which implies that $n^2 = 63 - \sqrt{x}$. This further simplifies to $\sqrt{x} = 63 - n^2$, and squaring both sides gives us $x = (63-n^2)^2$.
    \item (\textcolor{green}{Correct step}) The perfect squares less than or equal to $63$ are $1, 4, 9, 16, 25, 36, 49, 64$.
    \item (\textcolor{green}{Correct conclusion}) Therefore, the number of real values of $x$ for which $\sqrt{63-\sqrt{x}}$ is an integer is equal to the number of perfect squares less than or equal to $63$, which is $8$.
  }
\end{minipage}

\vspace{-4pt}
\caption{\textbf{Qualitative comparison.} CTRLS avoids hallucinated symbolic reasoning and correctly counts the number of valid values based on integer square analysis, while the baseline follows an incorrect symbolic path.}

\label{fig:qualitative3}
\end{figure}